\documentclass[a4paper,conference]{IEEEtran}
\IEEEoverridecommandlockouts
\usepackage{cite}
\usepackage{amsmath,amssymb,amsfonts}
\usepackage{algorithm}
\usepackage{algpseudocode}
\usepackage{graphicx}
\usepackage{textcomp}
\usepackage{xcolor}
\usepackage{booktabs}
\usepackage{multirow}
\usepackage{array}
\usepackage{newtxtext}
\usepackage{newtxmath}
\def\BibTeX{{\rm B\kern-.05em{\sc i\kern-.025em b}\kern-.08em
    T\kern-.1667em\lower.7ex\hbox{E}\kern-.125emX}}
\begin{document}

\title{FedTether: Cross-Attack Robust Aggregation\\
for Federated Learning on IoT Edge Networks}

\author{
\IEEEauthorblockN{Eyhab Al-Masri}
\IEEEauthorblockA{
\textit{School of Engineering and Technology} \\
\textit{University of Washington, Tacoma, WA, USA} \\
ealmasri@uw.edu
}
}

\maketitle

\begin{abstract}
Federated learning (FL) lets resource-constrained IoT and edge devices train a shared model without exchanging raw data, but compromised or faulty devices can poison the global model. Existing robust aggregators each fail in some regime: naive averaging (FedAvg) collapses under high-magnitude attacks, whereas distance- and statistics-based defenses (Krum, median, trimmed mean) sacrifice accuracy on non-IID data and remain vulnerable to stealthy attacks. We present FedTether, which reweights client updates by their deviation from a coordinate-wise median through a robust, dead-zoned score applied as a server-side optimizer step. Across three datasets, three attack types, and Byzantine fractions up to 40\%, FedTether is the only method top-tier in every regime: it matches FedAvg under stealthy attacks with no robustness tax, exceeds robust baselines by up to 42 accuracy points, and stays accurate where high-magnitude attacks reduce FedAvg to near-random. These properties persist at 50 clients under partial participation, with no shared data or validation set.
\end{abstract}


\section{Introduction}
The proliferation of smartphones, wearables, smart-home appliances, and industrial sensors has shifted computation toward the network edge. Federated learning (FL)~\cite{mcmahan2017} enables such IoT and edge devices to collaboratively train a shared model by exchanging only model updates rather than raw data, mitigating the privacy and bandwidth costs of centralized data collection. In the canonical algorithm, FedAvg, the server aggregates client updates as a sample-size-weighted average and broadcasts the resulting model to the devices.

This distributed structure is also a vulnerability. Edge and IoT devices are inexpensive, accessible, and infrequently patched, which makes them realistic targets for compromise; even absent an adversary, hardware faults and transmission errors corrupt updates. Such Byzantine participants can substantially degrade the global model~\cite{blanchard2017}. Data heterogeneity compounds this: each device holds a distinct, non-IID partition; honest updates thus vary widely. An effective defense must separate malicious deviation from benign heterogeneity.

Many robust aggregation rules have been proposed, including distance-based selection (Krum and Multi-Krum)~\cite{blanchard2017}, the coordinate-wise median, and the trimmed mean~\cite{yin2018}, yet none proves dependable across the conditions an edge deployment encounters. Our experiments confirm three failure modes: undefended averaging is accurate under mild attacks but collapses under high-magnitude poisoning; distance-based methods such as Multi-Krum withstand high-magnitude attacks yet fail under stealthy ones and strong non-IID heterogeneity; and coordinate-wise statistics incur a substantial penalty on non-IID data and can themselves collapse. Because the attack type cannot be anticipated, single-regime robustness has limited practical value.

FedAvg alone is insufficient, despite its strong performance under benign and stealthy conditions. It offers no protection against high-magnitude faults: under a Gaussian or sign-flipping attack---a malfunctioning sensor or corrupted transmission---it degrades to near-random accuracy (Section~\ref{sec:results}). Because such faults are unavoidable in practice, a single occurrence may forfeit the entire model.

We propose FedTether, which stays accurate across attack types and under non-IID heterogeneity yet incurs no penalty absent an attack. FedTether scores each client update by its squared distance to the coordinate-wise median, normalizes the scores with a floored, MAD-based scale, and applies a \emph{dead-zone} that down-weights only clear outliers before combining the survivors as a server-side optimizer step in delta space. The dead-zone is key: when no update is anomalous, all weights are unchanged and FedTether reduces \emph{exactly} to FedAvg, paying no robustness tax. Our contributions are as follows:
\begin{itemize}

\item We introduce FedTether, a median-referenced, MAD-scored, dead-zoned aggregation rule applied as a server-side optimizer step, requiring no shared data or validation set.
\item We show cross-attack robustness on CIFAR-10 (stealthy ALIE, sign-flipping, Gaussian): every baseline---including the closely related RFA---fails under some attack, whereas FedTether stays top-tier.
\item We quantify the robustness tax across MNIST, CIFAR-10, and CIFAR-100 under growing Byzantine fractions, and at 50 clients under partial participation.
\item We report five-seed results with paired-bootstrap significance, identifying when the fixed and adaptive variants are statistically equivalent.
\end{itemize}

\section{Related Work}
\textbf{Robust aggregation under heterogeneity.} Byzantine-robust aggregation replaces the FedAvg mean with a rule that tolerates a minority of corrupted updates (Table~\ref{tab:relwork}). Krum and Multi-Krum~\cite{blanchard2017} retain the updates closest to their neighbors; the coordinate-wise median and trimmed mean~\cite{yin2018} discard per-coordinate extremes; Bulyan~\cite{elmhamdi2018} composes the two, and Zeno~\cite{xie2019} scores updates by a loss-derived suspicion measure. Devised for the IID setting, these rules treat distant updates as malicious. Under the non-IID partitions typical of edge deployments~\cite{kairouz2021,li2020}, however, honest clients are themselves distant; distance- and rank-based filtering thus discards benign heterogeneity and incurs a standing accuracy penalty---the robustness tax---that optimization-side methods such as SCAFFOLD~\cite{scaffold} leave unaddressed for adversarial filtering.

\textbf{Attacks.} These defenses are not uniformly effective. Stealthy, variance-aware perturbations (ALIE)~\cite{baruch2019} remain within the honest spread and evade distance-based filters, whereas optimization-based model poisoning~\cite{fang2020,bhagoji2019} and backdoor attacks~\cite{bagdasaryan2020} tailor updates to a specific defense. Because the attack family is unknown in advance, a defense must hold across stealthy and high-magnitude regimes simultaneously---the cross-attack requirement that motivates our evaluation.

\textbf{Trust-, history-, and geometry-based defenses.} A second line reduces reliance on raw distance. FLTrust~\cite{fltrust} bootstraps trust from a small server-held root dataset, an assumption unavailable in our threat model. History-based centered clipping~\cite{karimireddy2021} and the geometric median (RFA)~\cite{pillutla} reweight or clip every client each round using only the submitted updates, as does FedTether; backdoor-specific filters~\cite{flame} target a complementary threat. The distinguishing mechanism is the dead-zone: clipping and RFA apply a continuous, always-on contraction that never fully disengages; they therefore down-weight honest non-IID clients even absent an attack and retain a residual tax, while selection rules such as Krum discard a fixed number of clients regardless of whether any are malicious. FedTether instead disengages entirely---reducing exactly to FedAvg---whenever no update crosses the dead-zone, eliminating the tax while requiring neither a root dataset nor the attacker count that Multi-Krum assumes.

\begin{table}[t]

\caption{Comparison of aggregation methods.
Non-IID indicates the ability to operate effectively
under heterogeneous client data, rather than guaranteed
immunity to client drift.}
\label{tab:relwork}
\centering
\renewcommand{\arraystretch}{1.2}
\setlength{\tabcolsep}{2pt}
\resizebox{\columnwidth}{!}{%
\begin{tabular}{lccccc}
\toprule
\textbf{Method} & \textbf{Non-IID} & \textbf{Stealthy} & \textbf{High-mag.}
& \textbf{No tax} & \textbf{No shared} \\
 & & \textbf{attack} & \textbf{attack} & & \textbf{data} \\
\midrule

FedAvg \cite{mcmahan2017} & \checkmark & \checkmark & $\times$ & \checkmark & \checkmark \\
Median \cite{yin2018}          & $\times$  & $\times$   & \checkmark & $\times$   & \checkmark \\
Trimmed mean \cite{yin2018}    & $\times$  & $\times$   & $\times$   & $\times$   & \checkmark \\
Multi-Krum \cite{blanchard2017}& $\times$  & $\times$   & \checkmark & $\times$   & \checkmark \\
RFA \cite{pillutla}            & \checkmark & $\times$   & \checkmark & $\times$   & \checkmark \\
FLTrust \cite{fltrust}         & \checkmark & \checkmark & \checkmark & $\times$   & $\times$   \\
\textbf{FedTether (ours)}      & \checkmark & \checkmark & \checkmark & \checkmark & \checkmark \\
\bottomrule
\end{tabular}%
}
\end{table}

\section{Threat Model and Method}\label{sec:threat}

\subsection{Federated Setting and Threat Model} We consider $N$ edge clients and one server, with a subset of clients participating in each communication round. Our primary experiments assume full participation, where all $N$ clients contribute each round, while the scalability experiments additionally consider partial participation with randomly sampled clients. In round $t$, the server broadcasts the global model $\theta^t$ to participating clients, which train locally on their private non-IID data and return their updates. A fixed fraction of the overall client population is \emph{Byzantine} and may return arbitrary updates intended to corrupt the global model when selected to participate. We assume an honest server with no access to raw client data and no clean validation set, which reflects the realistic IoT setting. We study three attack types representative of distinct edge failure modes:

\begin{itemize}
\item \emph{Stealthy (ALIE-style)}~\cite{baruch2019}: a small, hard-to-detect perturbation along the aggregate's variance direction, modeling a sophisticated intruder. As implemented, it is single-client and non-adaptive: attackers do not coordinate, observe other updates, or optimize against the defense.
\item \emph{Sign-flipping}: the attacker negates its honest update, modeling a malfunctioning or inverted node.
\item \emph{Gaussian}: the update is replaced with large Gaussian noise, modeling a faulty sensor or transmission.
\end{itemize}

\subsection{FedTether Aggregation}
\begin{figure}[t]
\centering
\includegraphics[width=\columnwidth]{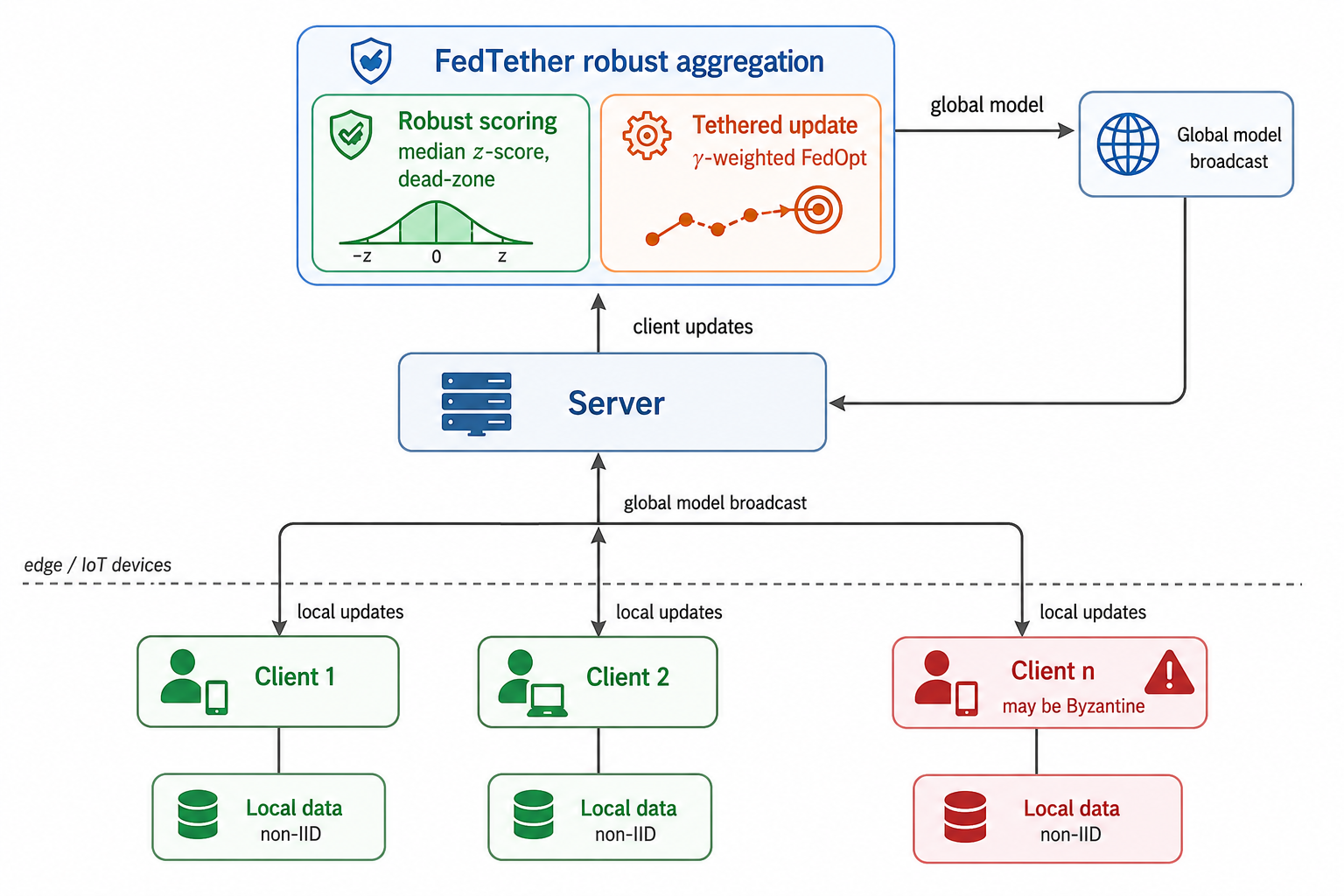}
\caption{FedTether architecture: median-referenced robust scoring and a $\gamma$-weighted server optimizer step.}

\label{fig:arch}
\end{figure}

Fig.~\ref{fig:arch} shows the overall system architecture. FedTether operates entirely on model updates in \emph{delta space}, maintaining a running global model and applying a robustly weighted aggregate delta as an optimizer step.

Let $\theta^t$ be the running global model and $c_i$ the parameters returned by client $i$ with $n_i$ local samples. The per-client update is $\delta_i = c_i - \theta^t$. FedTether adopts the coordinate-wise median of the updates as a robust reference direction, since the median is insensitive to a minority of arbitrary outliers:
\begin{equation}
m = \operatorname{median}\big(\{\delta_i\}_{i=1}^{N}\big)\quad\text{(coordinate-wise)} .
\end{equation}
Each update's deviation from this reference is summarized by a scalar energy $e_i = \lVert \delta_i - m \rVert_2^2$. To identify anomalous energies without being misled by the wide but benign spread of non-IID clients, we standardize $e_i$ with a robust location and scale derived from the median and the median absolute deviation (MAD):
\begin{align}
\mathrm{MAD} &= 1.4826 \cdot \operatorname{median}\big(\lvert e_i - \operatorname{median}(e) \rvert\big), \\
s &= \max\!\big(\mathrm{MAD},\, \rho \cdot \operatorname{median}(e)\big) + \varepsilon, \\
z_i &= \big(e_i - \operatorname{median}(e)\big)/s .
\end{align}
The scale floor $\rho\,\operatorname{median}(e)$ prevents the MAD from collapsing toward zero at small client counts (which would cause $z_i$ to diverge), stabilizing the score at the client counts typical of cross-silo and edge FL.

A \emph{dead-zone} of width $\tau$ ensures that only clearly anomalous updates are penalized, leaving the normal spread of honest non-IID clients unaffected. Each client's soft weight is
\begin{equation}
\gamma_i = \exp\!\big(-\beta \cdot \max(0,\, z_i - \tau)\big),
\label{eq:gamma}
\end{equation}
and the aggregation weights combine this factor with the sample count, $w_i = \gamma_i n_i / \sum_{j} \gamma_j n_j$. When no update exceeds the dead-zone, $\gamma_i = 1$ for all $i$, the weights reduce to the sample-size weights, and aggregation reduces \emph{exactly} to FedAvg; FedTether therefore adds no accuracy penalty when there is nothing to defend against. The aggregate delta is applied as a server-side optimizer step with learning rate $\eta$ (and optional momentum, disabled in our experiments):
\begin{equation}
\theta^{t+1} = \theta^{t} + \eta \sum_i w_i\, \delta_i .
\end{equation}

\subsection{Static and Adaptive $\beta$}
The sharpness $\beta$ in~\eqref{eq:gamma} governs how aggressively outliers are suppressed. FedTether-\textbf{Static} uses a fixed $\beta$. FedTether-\textbf{Adaptive} sets $\beta$ from the observed threat level via the maximum score in the round,
\begin{equation}
\beta = \beta_{\min} + (\beta_{\max}-\beta_{\min})\,
\sigma\!\big(\kappa\,(\max_i z_i - z_0)\big),
\end{equation}
where $\sigma$ is the logistic function; the schedule raises $\beta$ when a strong outlier is present and relaxes it otherwise. Algorithm~\ref{alg:fedtether} summarizes one round.

\begin{algorithm}[t]
\caption{FedTether aggregation (one round)}
\label{alg:fedtether}
\begin{algorithmic}[1]
\Require global model $\theta^t$; client params $\{c_i\}$, sizes $\{n_i\}$;
dead-zone $\tau$; floor $\rho$; server lr $\eta$
\State $\delta_i \gets c_i - \theta^t$ \textbf{ for all } $i \in \{1,\dots,N\}$
\State $m \gets \operatorname{median}_{\text{coord}}(\{\delta_i\})$
\State $e_i \gets \lVert \delta_i - m \rVert_2^2$
\State $\mathrm{MAD} \gets 1.4826\,\operatorname{median}(\lvert e - \operatorname{median}(e)\rvert)$
\State $s \gets \max(\mathrm{MAD}, \rho\,\operatorname{median}(e)) + \varepsilon$
\State $z_i \gets (e_i - \operatorname{median}(e))/s$
\State choose $\beta$ (fixed, or adaptive from $\max_i z_i$)
\State $\gamma_i \gets \exp(-\beta\max(0, z_i-\tau))$
\State $w_i \gets \gamma_i n_i / \sum_j \gamma_j n_j$
\State $\theta^{t+1} \gets \theta^t + \eta \sum_i w_i \delta_i$
\State \Return $\theta^{t+1}$
\end{algorithmic}
\end{algorithm}

\section{Experimental Setup}
We implement all methods in Flower with $N=10$ fully participating clients and non-IID data partitioned by a Dirichlet distribution ($\alpha=0.3$). The model is a compact convolutional network with GroupNorm, since BatchNorm statistics are corrupted under non-IID averaging. Clients train for 2 local epochs (SGD; lr $0.01$; momentum $0.9$; batch size $32$). We run MNIST, CIFAR-10, and CIFAR-100 for 15, 25, and 30 rounds, each over 5 seeds with identical per-seed initialization, reporting final-round mean and standard deviation.

We compare FedAvg, Median, Trimmed Mean, Multi-Krum, RFA (geometric median via Weiszfeld iteration)~\cite{pillutla}, and FedTether (Static and Adaptive). RFA is evaluated on the cross-attack configuration and the no-attack point; Multi-Krum receives the \emph{true} attacker count, an oracle advantage FedTether lacks. FedTether's hyperparameters are frozen study-wide: $\tau=3.0$, $\rho=0.5$, $\varepsilon=10^{-12}$, $\eta=1.0$ (no momentum), static $\beta=2.0$, and adaptive $\beta\in[0.2,4.0]$ with $\kappa=1.5$, $z_0=3.0$.

\section{Results}
\label{sec:results}

\subsection{Cross-Attack Robustness}
\begin{figure}[t]
\centering
\includegraphics[width=1.0\columnwidth]{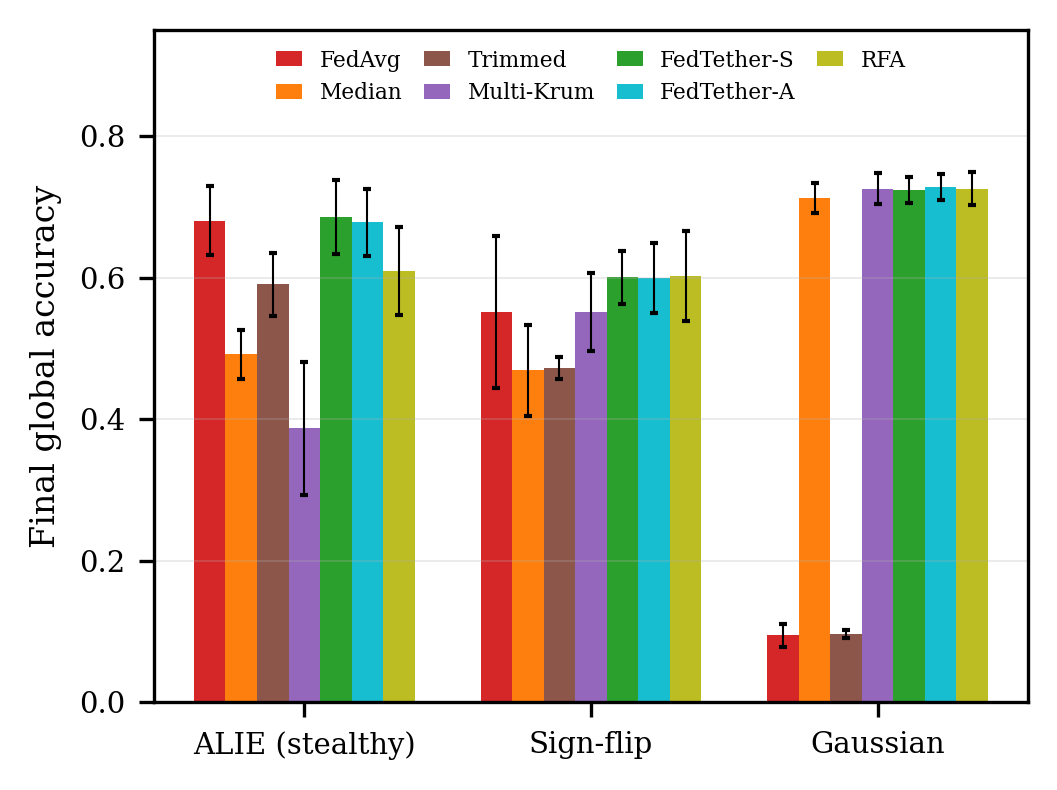}
\caption{CIFAR-10 final accuracy, 30\% Byzantine: every baseline collapses somewhere; FedTether does not.}
\label{fig:crossattack}
\end{figure}

Table~\ref{tab:crossattack} and Fig.~\ref{fig:crossattack} report CIFAR-10 final accuracy at 3 Byzantine clients (30\%) under all three attacks---the only configuration in which the attack type is varied, and thus the basis for the cross-attack claim. The pattern is consistent: every baseline fails under at least one attack. Multi-Krum, strong under the loud attacks, collapses under the stealthy ALIE attack ($0.387$). FedAvg and Trimmed Mean, competitive under the stealthy attack, fall to near-random accuracy under Gaussian noise ($0.094$ and $0.096$). FedTether (Static and Adaptive) is the only method that remains top-tier under all three attacks, leading or matching the best baseline in every column; FedAvg is competitive only in the stealthy and no-attack regimes, where FedTether equals it.

RFA, the method most closely related to FedTether, likewise reweights all clients using only the submitted updates, without a trusted dataset or known attacker count. It matches FedTether within noise under both loud attacks (sign-flip, $0.602$ vs.\ $0.600$; Gaussian, $0.726$ vs.\ $0.724$), but under the stealthy ALIE attack degrades to $0.609$---below both FedTether ($0.685$; paired-bootstrap $95\%$ CI $[+0.049,+0.104]$) and FedAvg ($0.681$). Its continuous, always-on reweighting penalizes honest but distant clients precisely when the attack hides within the benign non-IID spread, the failure the dead-zone avoids.

\begin{table}[t]
\caption{CIFAR-10 final accuracy at 30\% Byzantine under three attacks. Best per column in \textbf{bold}.}
\label{tab:crossattack}
\centering
\renewcommand{\arraystretch}{1.2}
\setlength{\tabcolsep}{3.5pt}
\begin{tabular}{lccc}
\toprule
\textbf{Method} & \textbf{ALIE (stealthy)} & \textbf{Sign-flip} & \textbf{Gaussian} \\
\midrule
FedAvg          & $0.681 \pm .049$ & $0.551 \pm .107$ & $0.094 \pm .016$ \\
Median          & $0.492 \pm .034$ & $0.469 \pm .064$ & $0.713 \pm .022$ \\
Trimmed mean    & $0.590 \pm .045$ & $0.472 \pm .015$ & $0.096 \pm .006$ \\
Multi-Krum      & $0.387 \pm .094$ & $0.552 \pm .055$ & $0.726 \pm .021$ \\
RFA \cite{pillutla} & $0.609 \pm .062$ & $\mathbf{0.602 \pm .064}$ & $0.726 \pm .024$ \\
FedTether-S     & $\mathbf{0.685 \pm .052}$ & $0.600 \pm .037$ & $0.724 \pm .019$ \\
FedTether-A     & $0.678 \pm .048$ & $0.600 \pm .050$ & $\mathbf{0.728 \pm .018}$ \\
\bottomrule
\end{tabular}
\end{table}

\begin{figure}[t]
\centering
\includegraphics[width=\columnwidth]{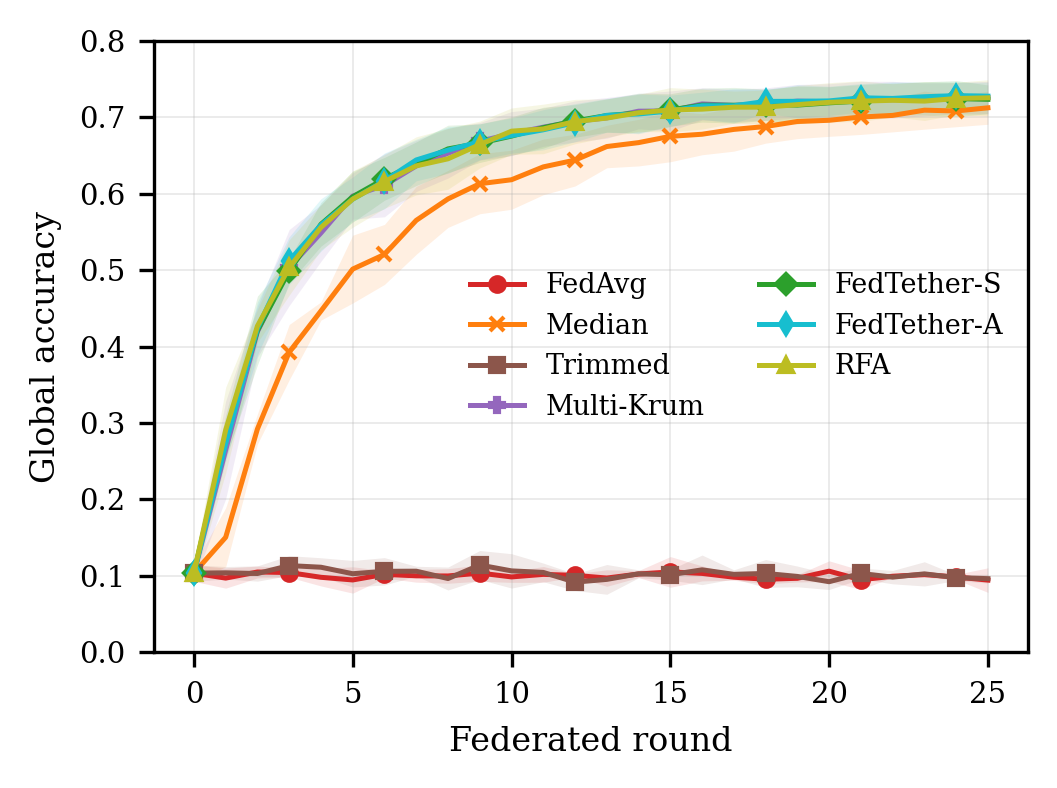}
\caption{CIFAR-10 over rounds, Gaussian attack (30\% Byzantine). FedAvg/Trimmed stay near-random; others recover.}
\label{fig:gaussian}
\end{figure}

The Gaussian case is the most pronounced. Fig.~\ref{fig:gaussian} shows accuracy over rounds: FedAvg and Trimmed Mean never leave random-guess accuracy ($\approx0.10$), because large-magnitude updates dominate the average and the symmetric trim, whereas FedTether's median reference and dead-zone isolate and suppress them, reaching $0.72$---a margin of more than 60 points over undefended averaging. This result concretely answers why undefended averaging is unsafe for edge deployment: a single class of loud fault renders it non-functional.

\subsection{Robustness Tax and Degradation}
Varying the number of Byzantine clients from 0 to 4 on CIFAR-10 under the stealthy ALIE attack, FedTether tracks undefended FedAvg almost exactly, confirming that it pays no robustness tax under non-IID data, while the robust baselines diverge downward as the attacker count grows. At 4 attackers (40\%), Multi-Krum and Median fall to $0.148$ and $0.318$, whereas FedTether (Static) retains $0.566$---a margin of up to 42 points. This behavior holds across task difficulty (Table~\ref{tab:crossdataset}): under the same attack at 3 attackers, FedTether matches FedAvg on CIFAR-100 ($0.353$ vs.\ $0.355$) while Median and Multi-Krum fall to $0.234$ and $0.202$; on MNIST all methods sit near the saturated ceiling.

\begin{table}[t]
\caption{Cross-dataset accuracy, stealthy ALIE (30\% Byzantine); FedAvg is the no-tax reference; best robust method per column in \textbf{bold}.}
\label{tab:crossdataset}
\centering
\renewcommand{\arraystretch}{1.2}
\setlength{\tabcolsep}{3pt}
\begin{tabular}{lccc}
\toprule
\textbf{Method} & \textbf{MNIST} & \textbf{CIFAR-10} & \textbf{CIFAR-100} \\
\midrule
FedAvg        & $0.987 \pm .003$ & $0.681 \pm .049$ & $0.355 \pm .027$ \\
Median        & $0.973 \pm .006$ & $0.492 \pm .034$ & $0.234 \pm .011$ \\
Trimmed mean  & $\mathbf{0.981 \pm .004}$ & $0.590 \pm .045$ & $0.288 \pm .014$ \\
Multi-Krum    & $0.854 \pm .103$ & $0.387 \pm .094$ & $0.202 \pm .031$ \\
FedTether-S   & $\mathbf{0.981 \pm .010}$ & $\mathbf{0.685 \pm .052}$ & $0.353 \pm .023$ \\
FedTether-A   & $0.980 \pm .008$ & $0.678 \pm .048$ & $\mathbf{0.354 \pm .029}$ \\
\bottomrule
\end{tabular}
\end{table}

\subsection{Scalability and Partial Participation}
\begin{figure}[t]
\centering
\includegraphics[width=1.0\columnwidth]{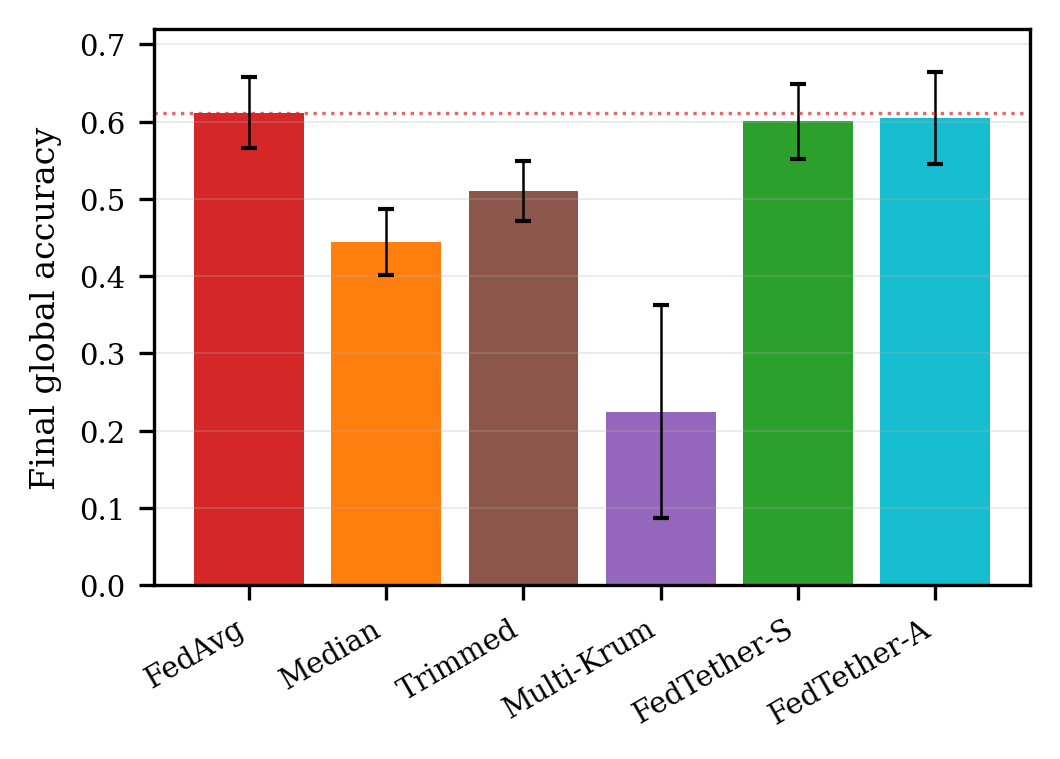}
\caption{50 clients, 40\% participation, 30\% Byzantine (ALIE). FedTether tracks FedAvg; baselines fall below.}
\label{fig:scale}
\end{figure}

To confirm these findings are not an artifact of the small-client setup, we ran CIFAR-10 with 50 clients under stealthy ALIE, 40\% partial participation (20 sampled per round) modeling intermittent edge connectivity, and 15 Byzantine clients (30\%). Under this dynamic sampling---a different non-IID subset and fluctuating attacker count each round---Multi-Krum collapsed to $0.225 \pm .138$, its variance reflecting instability as the sampled attacker count varies. Undefended averaging reached $0.612 \pm .047$, and FedTether (Static) tracked it at $0.601 \pm .049$, while Median and Trimmed mean lagged at $0.444$ and $0.510$ (Fig.~\ref{fig:scale}). FedTether's median-referenced scoring and floored scale thus remain stable at scale, paying no robustness tax.

\subsection{Static versus Adaptive}
FedTether-Static and FedTether-Adaptive perform within noise of each other across all conditions (e.g., $0.600$ vs.\ $0.600$ under sign-flip; $0.724$ vs.\ $0.728$ under Gaussian). A fixed $\beta$ therefore suffices in every regime tested here, and we report the adaptive variant for completeness without claiming an advantage. Its potential benefit under time-varying attacker intensity is outside the scope of this study.

\subsection{Discussion}
Two mechanisms explain this consistent behavior. Under high-magnitude attacks (sign-flipping, Gaussian), malicious updates lie far from the coordinate-wise median, yielding large energies $e_i$ and scores $z_i$; the exponential weight in~\eqref{eq:gamma} drives their contribution toward zero. FedTether thus stays accurate under Gaussian noise, whereas undefended averaging and the symmetric trimmed mean still incorporate the large updates and collapse to near-random accuracy. Conversely, the dead-zone $\tau$ leaves the wide but benign spread of honest non-IID clients unpenalized: absent a clear outlier, all weights remain unity and aggregation reduces to FedAvg. FedTether therefore pays no robustness tax across increasingly difficult datasets, whereas distance- and statistics-based baselines down-weight honest heterogeneity and lose accuracy as difficulty grows.

The dead-zone also distinguishes FedTether from RFA, its closest relative: RFA's smoothed geometric median never disengages, admits no exact reduction to FedAvg, and down-weights honest but distant non-IID clients whenever an attack renders the update distribution asymmetric---hence it matches FedTether under high-magnitude attacks but degrades under stealthy ones. The baselines exhibit complementary weaknesses: FedAvg fails catastrophically under high-magnitude faults, while distance- and statistics-based defenses pay a standing penalty on non-IID data and remain fragile under stealthy attacks. FedTether reconciles these regimes---the median reference suppresses outliers, the dead-zone preserves benign heterogeneity---sustaining top-tier accuracy across all regimes without prior knowledge of the attack type.

A remaining question is whether an adversary can exploit the dead-zone directly. The attacks evaluated here are non-adaptive (Section~\ref{sec:threat}); a defense-aware adversary would instead keep its robust score within the dead-zone, $z_i \le \tau$, to evade suppression, but the score's construction bounds the damage. Since $z_i$ standardizes the energy $e_i = \lVert \delta_i - m \rVert_2^2$ against the robust spread $s$ of honest updates, any update with $z_i \le \tau$ is energetically comparable to an atypical honest client, and its influence is bounded by the heterogeneity the sample-weighted average already absorbs; an update large enough to move the model instead raises $z_i$ above $\tau$ and is suppressed by~\eqref{eq:gamma}. The adversary thus faces a dilemma: bounded harm within the dead-zone, or suppression. This analysis concerns a single, non-colluding client; adaptive and colluding adversaries---particularly coordinated attacks that displace the median---remain future work.

\section{Conclusion}

We presented FedTether, a robust aggregation rule for FL
on heterogeneous IoT edge networks. Using a coordinate-wise
median, a floored MAD-based scale, and dead-zoned reweighting,
FedTether maintains competitive accuracy across stealthy
and high-magnitude attacks and heterogeneous datasets,
while reducing exactly to FedAvg when no update exceeds
the dead-zone. It is the only method robust across all
tested attack regimes, preventing the near-random accuracy
observed with undefended averaging under high-magnitude
attacks. Its effectiveness also extends to larger client
populations under partial participation, without requiring
shared data or a trusted validation set. Future work
includes adaptive and colluding adversaries, scaling to
hundreds of physical devices with stragglers, and convergence
analysis of dead-zoned aggregation.

\end{document}